\documentclass[
  manuscript=article-type,
  year=2026,
]{cup-journal}

\usepackage{amsmath}
\usepackage[nopatch]{microtype}
\usepackage{booktabs}

\usepackage{latexsym}
\usepackage{amssymb}
\usepackage{amsmath}
\usepackage{amsthm}
\usepackage{booktabs}
\usepackage{enumitem}
\usepackage{graphicx}
\usepackage{color}
\usepackage{booktabs}
\usepackage{float}
\usepackage{amsmath}
\usepackage{algorithm}
\usepackage{algpseudocode}
\usepackage{array}
\usepackage{multirow}
\usepackage{adjustbox}
\usepackage{graphicx}
\usepackage{caption}
\usepackage{colortbl}
\usepackage{arydshln}
\usepackage[most]{tcolorbox}
\usepackage{placeins}
\usepackage{times}
\usepackage{latexsym}
\usepackage{todonotes}
\usepackage{xcolor}
\usepackage{enumitem}
\usepackage{soul}
\usepackage{longtable}
\usepackage{url}
\usepackage{hyperref}
\usepackage{wrapfig} 
\usepackage{dashrule}
\usepackage[normalem]{ulem}
\usepackage{tabularray}
\usepackage{orcidlink}

\title{Controlled Attribute-Specific Summarization of Interrogative Dialogues}

\author{A Aditya Bhardwaj \orcidlink{0009-0003-1336-6597}}
\affiliation{Department of Computer Science and Engineering, IIIT Delhi, New Delhi, 110020, Delhi, India}

\author{Arjit Singh Arora}
\affiliation{Department of Computer Science and Engineering, IIIT Delhi, New Delhi, 110020, Delhi, India}

\author{Md Shad Akhtar \orcidlink{0000-0002-2033-2382}}
\affiliation{Department of Computer Science and Engineering, IIIT Delhi, New Delhi, 110020, Delhi, India}
\email[F. Author]{shad.akhtar@iiitd.ac.in}


\begin{document}

\begin{abstract}
Effective summarization of interrogative dialogues is a critical task in forensic and investigative settings, requiring high factual accuracy, coherence, and attribute-specific relevance. In this work, we introduce \textbf{CASPER}, a novel Chain-of-Thought Attribute-Specific Prompting for Evaluative Summarization framework that leverages structured prompting and iterative refinement to generate high-quality summaries of interrogator-witness interactions. We construct \textbf{MINDSum}, a dataset extending the \textbf{MIND} corpus, comprising 6,000 utterance pairs annotated with event details, factual statements, character descriptions, and fillers. CASPER employs \textbf{RoleEval}, a hierarchical evaluation mechanism where multiple roles (\textit{officer}, \textit{inspector}, \textit{senior inspector}) iteratively assess summaries based on predefined criteria. By integrating entity extraction and structured feedback loops, CASPER significantly improves factual consistency and contextual completeness compared to existing baselines. Experimental results demonstrate that our framework outperforms standard summarization models on both lexical (ROUGE) and semantic (BERTScore) metrics, while human evaluation confirms its alignment with expert reasoning. Our findings underscore the potential of controlled summarization in high-stakes domains, paving the way for AI-driven forensic intelligence.
\end{abstract}

\section{Introduction}
Witness testimonies play a critical role in criminal investigations, often serving as primary narrative evidence when physical traces are incomplete or ambiguous. However, interrogation transcripts are typically verbose, fragmented, and cognitively disorganized, containing redundancies, inconsistencies, and peripheral details. In high-stakes forensic settings, producing accurate summaries is not merely a convenience—it is a decision-support necessity. Errors in summarization may distort investigative reasoning, misrepresent intent, or omit legally salient facts.

\begin{figure}[t]
  \centering
  \includegraphics[width=0.8\columnwidth]{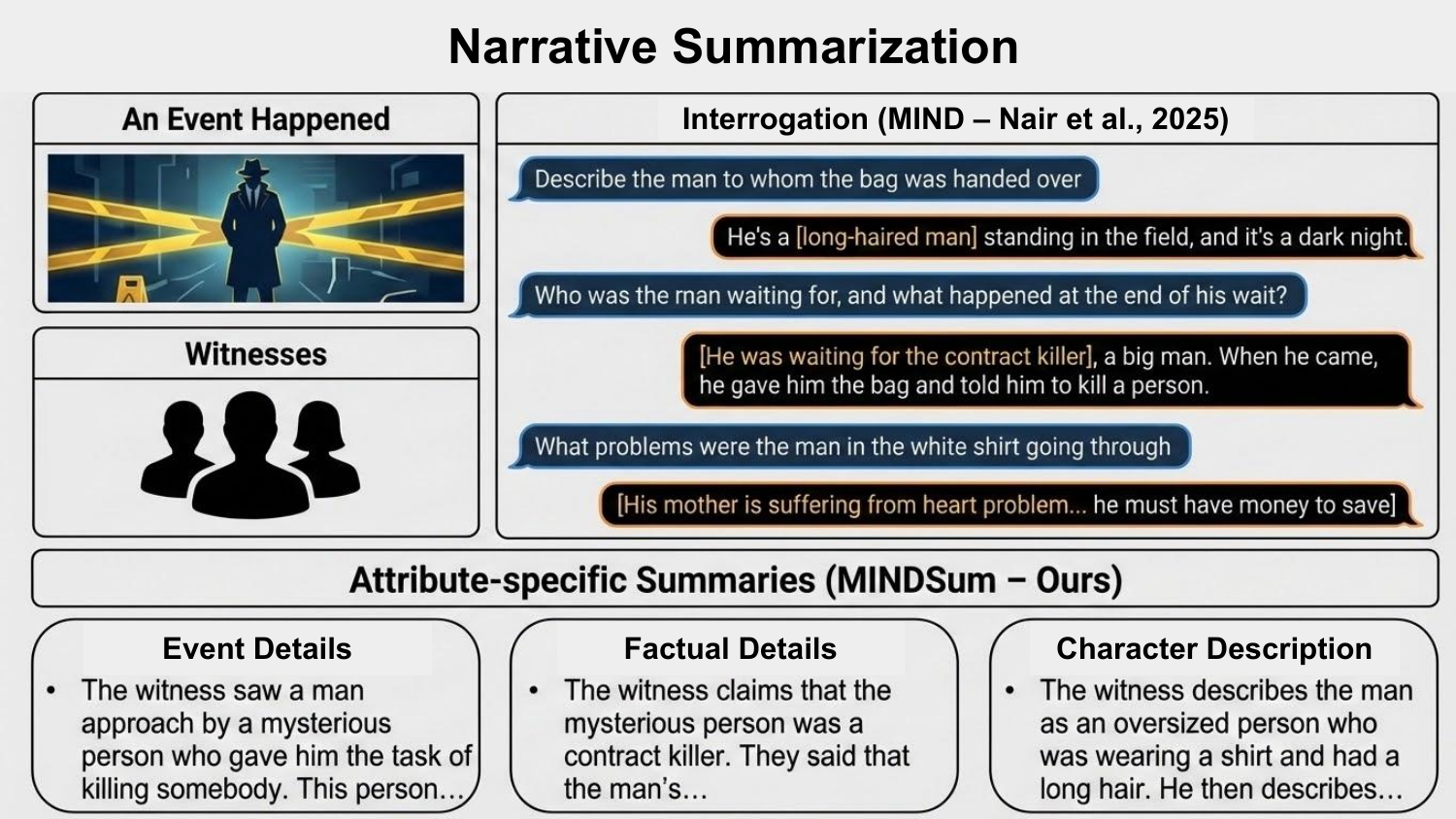}
  \caption{Overview of the witness interrogation and summarization process. Witness testimonies, often verbose and inconsistent, are summarized based on key attributes, event details, factual claims, and character descriptions, to ensure clarity and forensic relevance.\protect\footnotemark}
  \label{fig:problemOverview}
\end{figure}

Figure \ref{fig:problemOverview} depicts how legal professionals gather testimonies that, while detailed, often contain inconsistencies, ambiguities, and extraneous details. After collecting witness testimonies, legal professionals prepare a concise overview known as a deposition summary. This summary condenses extensive transcripts into a more digestible format, making it easier for law enforcers to reference as they prepare their cases. The most effective deposition summaries capture essential details, allowing readers to grasp the core components of the testimony in just a few minutes \citep{Feller_2024}. 

Research in legal psychology \citep{Loh81} emphasizes that reliable testimony evaluation depends on structured representation of narrative content—specifically, what occurred, what was factually asserted, and how individuals were described in relation to events. Prior empirical frameworks \citep{OTGAAR2024100562} argue that testimonial reliability hinges on correspondence between narrative structure and factual event reconstruction. 
Motivated by these findings, we propose a \textbf{controlled attribute-specific summarization} task that aims to distill complex narratives from investigative testimony into concise and relevant summaries. We operationalize forensic summarization through three investigative axes: event details, factual details, and character descriptions of subjects involved, which together capture the core informational structure necessary for legal assessment and can  provide a detailed and unambiguous scenario of the event to the law-and-enforcement agencies (LAEs). We conceive these attributes as follows:
\footnotetext{We use Gemini-nano-banana for aesthetically editing the images.\label{lbl}}

\begin{itemize}[leftmargin=*, itemsep=0pt]
    \item \textbf{Event Details:} This entails chronologically mapping out events, pinpointing when and where they occurred, and providing a clear understanding of the sequence of actions. Each event is supported by timestamped utterances, ensuring accuracy and reliability. For example, in Figure \ref{fig:problemOverview}, the witness describes how a man was approached by someone, capturing the "what" aspect.  
    \item \textbf{Factual Details:} In this category, objective claims are extracted and verified for consistency against accounts from other witnesses. The information is presented factually, highlighting essential truths without ambiguity. As shown in the Figure \ref{fig:problemOverview}, the witness claims that the person who approached the man was, in fact, a contract killer.
    \item \textbf{Character Descriptions:} Here, individuals involved in the event are profiled in a structured manner by detailing their attributes and roles in the events through witness account. This aspect ensures that all descriptions are consistent and provide a vivid picture of the people involved. In Figure \ref{fig:problemOverview}, the witness describes the man as an oversized person with long hair.
\end{itemize}

We aim to generate precise summaries for each category, preserving the integrity of the original testimony while preventing overlap between different summaries. This systematic approach helps investigators understand the full scope of the testimony, enabling informed decision-making and further analysis. To support this task, we introduce \textbf{MINDsum}, an extension of the \textbf{MIND} corpus (\textbf{M}ult\textbf{I}-eyewit\textbf{N}ess \textbf{D}eception) \citep{nair2025incongruenceidentificationeyewitnesstestimony}, enhanced with expert-annotated summaries aligned with three investigative attributes. The dataset consists of 5,785 annotated witness-interrogator dialogue pairs, focusing on forensic structured interrogations.

Despite advances in LLMs, existing summarization systems primarily optimize for surface-level fluency or lexical overlap, rather than structured investigative alignment. In forensic contexts, summaries must simultaneously satisfy multiple constraints: chronological coherence (event details), preservation of objective claims (factual details), and structured profiling of individuals (character descriptions). Current controllable summarization methods do not explicitly disentangle or enforce such multi-attribute constraints, nor do they provide hierarchical evaluation mechanisms aligned with investigative review processes. This creates a critical gap between general-purpose summarization and forensic-grade documentation.

To mitigate this, we propose \textbf{CASPER}, a \textbf{C}hain-of-Thought \textbf{A}ttribute-\textbf{S}pecific \textbf{P}rompting for \textbf{E}valuative Summa\textbf{R}ization framework. CASPER integrates attribute-guided prompting with iterative self-refinement to enforce structured forensic summarization. It uses the attributes defined above to decompose witness testimony summarization into three investigative axes: \textit{event details}, \textit{factual details}, and \textit{character description}. Each axis is governed by modular prompt templates that condition the LLM on forensic guidelines. This enables CASPER to manage conflicting constraints, a capability that has been underexplored in prior work \citep{russo-etal-2020-control}.

Moreover, we introduce \textbf{RoleEval}, a hierarchical evaluation framework that employs role-specific prompting to assess summary quality. RoleEval follows the structured approach of Promethee method \citep{mareschal2005promethee} to simulates forensic analysis by assigning distinct reviewer roles (e.g., \textit{officer}, \textit{inspector}, \textit{senior inspector}) to refine assessments based on investigative rubrics (e.g., factual completeness, coherence, attribute relevance). This layered approach ensures rigorous scrutiny, reducing biases and enhancing forensic summary quality. Our evaluation combines automatic and human assessments, incorporating lexical ROUGE \citep{lin-2004-rouge}, semantic BERTScore \citep{zhang2020bertscoreevaluatingtextgeneration}, and fine-grained attribute-specific evaluation of  \textbf{RoleEval}. Our results demonstrate that CASPER produces structured, forensic-grade summaries with high factual reliability.

\paragraph{Contribution} Our key contributions are summarized as follows:
\begin{itemize}[leftmargin=*, itemsep=0pt]
    \item We introduce \textbf{investigative summarization} as a controlled text-generation task that enforces constraints based on attributes such as event details, factual details, and character descriptions.
    \item We present \textbf{MINDsum}, a dataset for forensic summarization, offering detailed annotations for event details, factual details, and character descriptions
    \item We propose \textbf{CASPER}, a prompt-based framework integrating dynamic attribute-specific control summarization employing a hierarchical evaluation policy.
    \item We propose \textbf{RoleEval}, a hierarchical evaluation framework with role-specific assessment to measure faithfulness, factual consistency, and attribute alignment.
\end{itemize}

\paragraph{Reproducibility} The code and a subset of the dataset is available at \href{https://github.com/flamenlp/CASPER}{\url{www.github.com/flamenlp/CASPER}}. We are committed to releasing the full dataset and model checkpoints on acceptance. 

\section{Related Work}

Previous research on controlled summarization has primarily emphasized content planning through entity chains \citep{narayan-etal-2021-planning} and discrete control tokens \citep{he-etal-2022-ctrlsum}. More recent approaches incorporate coarse-grained constraints, such as length control and entity preservation \citep{tang-etal-2023-context}, as well as hybrid extractive-abstractive techniques, which collectively enhance the controllability of generated summaries. However, these methods often require extensive fine-tuning of pre-trained language models (PLMs) and remain limited in their ability to enforce multi-attribute constraints or structured forensic requirements. Moreover, conventional benchmarks primarily rely on lexical overlap metrics, e.g., ROUGE \citep{lin-2004-rouge}, which fail to capture key investigative properties such as temporal coherence, cross-witness consistency, and factual grounding \citep{kryscinski-etal-2019-neural, maynez-etal-2020-faithfulness}. These challenges underscore the need for a domain-specific framework that dynamically prioritizes investigative attributes while mitigating hallucination.

Controlled text summarization (CTS) enhances adaptability by incorporating user-specified constraints such as length, style, and content focus. Early models like CTRLsum \citep{he-etal-2022-ctrlsum} introduced control tokens, enabling parameter-efficient mechanisms for guiding summary generation. Recent advancements have refined CTS through discrete and continuous prompt tuning, leveraging pre-trained language models (PLMs). Techniques such as soft prompt tuning \citep{lester-etal-2021-power}, prefix-tuning \citep{li-liang-2021-prefix}, and hybrid approaches combining planning and guidance \citep{narayan-etal-2021-planning} have significantly improved control over summary attributes like conciseness and entity focus. \cite{Passali_Tsoumakas_2026} presents a technique for controllable abstractive summarization that utilizes various textual contexts, ranging from a brief text to multiple documents, to steer the emphasis of the resulting summary. It introduces an approach that employs a sentence BERT model to create embeddings that label keywords in the input document.

Prompt-based tuning has become a cornerstone of CTS, particularly in few-shot and parameter-efficient settings. For example, \citep{ravaut-etal-2023-promptsum} proposed PromptSum, which integrates entity-aware conditioning to enhance abstractive summarization. Chain-of-thought (CoT) prompting \citep{wei-etal-2022-chain} and instance-level attribute control \citep{liu-etal-2023-attribute} have further advanced content faithfulness, especially in dialogue summarization. 

Multi-attribute control frameworks have emerged to manage factors such as entity focus, discourse structure, and factual consistency \citep{urlana-etal-2024-controllable}. However, evaluating CTS remains challenging due to the limitations of existing metrics. Metrics like ROUGE \citep{lin-2004-rouge} and BERTScore \citep{zhang2020bertscoreevaluatingtextgeneration} primarily assess surface-level similarity, often failing to capture nuanced improvements in factual alignment or user alignment \citep{kryscinski-etal-2020-evaluating}. Recent advances in large language models (LLMs) and CoT prompting \citep{shi-etal-2022-multilingual} have improved interpretability and factual consistency, but challenges in efficiency and robustness persist. Large language models have significantly advanced conversational NLP tasks including dialogue generation, summarization, and chatbot systems \citep{SINGH2025100128}.

Interrogation-based summarization, which employs targeted prompts or questions to guide summary generation, shares strong conceptual ties with CoT prompting and self-refinement techniques like Tree-of-Thought (ToT) \citep{yao2023tree} and Reflexion \citep{shinn2023reflexion}. By framing summarization as a question-driven task (e.g., ``What are the main arguments in this document?''), interrogation-based methods structure the process to ensure alignment with user constraints, similar to the intermediate reasoning steps in CoT \citep{wei-etal-2022-chain}. For instance, \cite{wang-etal-2023-element} introduced Summary Chain-of-Thought (SumCoT), which decomposes summarization into steps such as identifying key elements and synthesizing them into a coherent summary. SumCoT achieves significant improvements, with ROUGE-L score gains of +4.33/+4.77 on datasets like CNN/DailyMail and BBC XSum, demonstrating the efficacy of structured reasoning in summarization.

Self-refinement methods like ToT \citep{yao2023tree} and Reflexion \citep{shinn2023reflexion} offer promising avenues for enhancing CTS. ToT extends CoT by exploring multiple reasoning paths in a tree-like structure, potentially enabling the generation of diverse candidate summaries that emphasize different aspects of the text (e.g., arguments, evidence, or conclusions). The model could evaluate these candidates based on coherence or user constraints, selecting the optimal summary. Reflection, which involves iterative self-assessment and refinement, could enable models to improve their summaries by critiquing initial outputs against metrics such as factual accuracy and conciseness. However, the application of ToT and Reflexion to summarization remains underexplored, and the current literature lacks specific examples in this domain, highlighting a research gap.

In contrast to prior controlled text summarization approaches that primarily focus on surface-level control signals such as length, entity prompts, or stylistic constraints, our work introduces a forensic-grounded, attribute-specific framework tailored to interrogative dialogues. While existing methods \citep{ravaut-etal-2023-promptsum, wang-etal-2023-element, naik-etal-2024-perspective, yao2023tree} leverage prompt tuning, chain-of-thought reasoning, or self-refinement to improve factuality and coherence, they do not explicitly decompose testimony into investigative axes such as event details, factual claims, and character descriptions, nor do they enforce strict attribute alignment in high-stakes domains. Moreover, previous evaluation strategies largely rely on lexical or semantic overlap metrics, which inadequately capture investigative relevance and forensic completeness. Our proposed CASPER framework advances the literature by integrating structured entity-guided prompting, iterative role-based refinement, and a hierarchical evaluation mechanism that mirrors real-world forensic review processes \citep{SHARMA2024100080}. This combination enables fine-grained control, improved factual robustness, and interpretable quality assessment, addressing key limitations in both controllable summarization and dialogue-based reasoning frameworks. 

\section{Data Preparation and Annotation Framework} \label{Dataset}

In this section, we introduce \textbf{MINDSum}, a novel dataset curated to support the summarization of interactions between an interrogator and a witness across the three forensic attributes. MINDSum extends the \textbf{MIND} (\textbf{M}ult\textbf{I}-eyewit\textbf{N}ess \textbf{D}eception) dataset \citep{nair2025incongruenceidentificationeyewitnesstestimony}, comprising $376$ interrogative testimonies against $149$ distinct crime-related events (e.g., robbery, theft, murder). These testimonies span more than $5,785$ utterance pairs between the interrogator and a witness. 

\begin{wrapfigure}{r}{7cm}
  \centering
  \includegraphics[width=\textwidth]{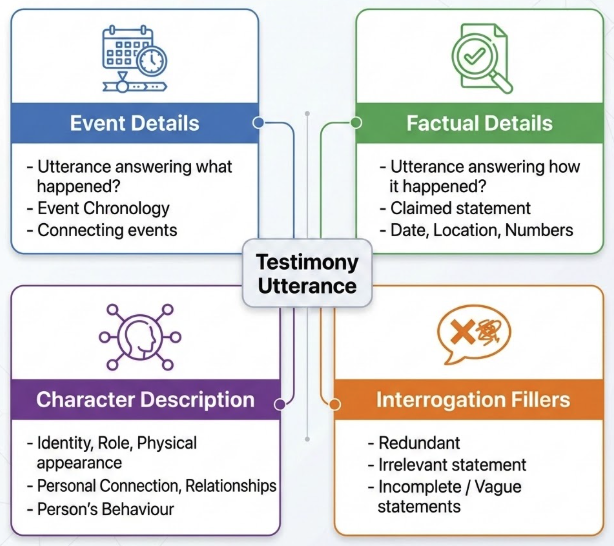}
  \caption{Annotation schema for MINDsum.\footref{lbl}}
  \label{fig:DialogueAnnotationTree}
\end{wrapfigure}

We extract testimonies from MIND \citep{nair2025incongruenceidentificationeyewitnesstestimony}, and manually summarize their narratives considering three attributes, i.e., \textit{event details}, \textit{factual details}, and \textit{character description}.

\subsection{Annotation}
For a narrative summary, it is important to accurately reflect the witness's original statements without assumptions or personal interpretations. To ensure this, MINDSum contains two-level of annotations: a) utterance-level attribute labels; and b) attribute-specific narrative summaries. At first, we tag a pair of interrogator-witness utterances as per the annotation schema depicted in Figure \ref{fig:DialogueAnnotationTree}. In addition to the three attributes defined earlier (i.e., ED, FD, and CD), we also use \textit{interrogation filler} (IF) label to filter non-essential, irrelevant, or redundant information considering the event narration. We follow a multilabel strategy, where an interrogator-witness utterance pair can take more than one attribute for overlapping cases. Subsequently, we annotate each testimony with three attribute-level summaries reflecting the \textit{event summary}, \textit{factual summary}, and \textit{character summary}.

\begin{table*}[t]
    \centering
    \renewcommand{\arraystretch}{1.3}
    \setlength{\tabcolsep}{6pt}
    \resizebox{1.0\textwidth}{!}{%
        \begin{tabular}{l p{14cm} c}
        \toprule
        {\bf Speaker} & {\bf Utterances} & {\bf Attr} \\
        \bottomrule

        \rowcolor{cyan!10} Interrogator & Describe the conversation between the boy and the chef. & \\
        \rowcolor{cyan!10} Witness & When the delivery boy arrived to pick up the order, the chef asked him, "Why are you always late? Your reviews and ratings are poor." The delivery boy responded, "Mind your own business. It’s not my problem." &  \multirow[c]{-2}{*}{ED} \\

        Interrogator & How was the behavior of the delivery boy towards the chef? &  \\
        Witness & He was angry at the chef because the chef said that his rating is not good. & \multirow[c]{-2}{*}{CD}\\

        \rowcolor{cyan!10}  Interrogator & Who was the delivery boy delivering the package to? Can you describe that person? &  \\
       \rowcolor{cyan!10}  Witness & Yeah, he was delivering the package to a woman. She was wearing a red blouse, and her hair was blonde. & \multirow[c]{-2}{*}{CD}\\

        Interrogator & So, how was the behavior of that woman, to whom the package was going to be delivered, towards the delivery boy? &  \\
         Witness & She was kind. She also gave him a tip of 400 dollars, which is a large amount. The boy was happy with the tip, and the lady also offered him a drink, as it was a party night. However, the boy declined, saying that the tip was enough for him. & \multirow[c]{-2}{*}{FD} \\

        \rowcolor{cyan!10} Interrogator & What was in the delivery package? &  \\
        \rowcolor{cyan!10}  Witness & The food that the lady ordered for the party. & \multirow[c]{-2}{*}{FD} \\

        Interrogator & So, after delivering the package to the woman, what happened? &  \\
        Witness & After delivering the package, the boy left the place, but as feedback, he got a message that said, "Please save me. I’m in danger." After thinking for some time, he went inside the room and saw that things were not in their right place. He got suspicious, and there was a noise coming from the inner room. So he went to the inner room and saw a man lying in the tub, and he was dead. After some time, he also saw the same lady, and the lady said, "You must not be here now. I will kill you." & \multirow[c]{-2}{*}{ED} \\

        \rowcolor{cyan!10}  Interrogator & What was written in the feedback that he got? &  \\
        \rowcolor{cyan!10}  Witness & Already told that it was written, "Please save me. I’m in danger." So he got a text message on his phone. & \multirow[c]{-2}{*}{IF} \\ \bottomrule
        \rowcolor{white} \multicolumn{3}{c}{\bf Attribute-specific Summaries} \\ \bottomrule
        \rowcolor{purple!15} ED Summary & The witness saw a delivery boy delivering a package to a woman at her house. After he delivered the package, he received a feedback from the which said, "Please save me. I'm in danger." & \\
        \rowcolor{green!10} FD Summary & The witness described the woman as kind towards the delivery boy when he arrived at the house, and she gave him a tip of 400 dollars, then offered him a drink, but the boy declined to drink. & \\
        \rowcolor{yellow!15} CD Summary & The witness describes that the delivery boy and the chef were not on good terms, and the boy's rating was also very low. They describe the woman as a young lady with blonde hair who was wearing a red blouse. & \\

        \bottomrule 
        \end{tabular}%
    }
    \caption{An Illustrative example of a witness testimony. The rightmost column shows the assigned attribute for each interrogator-witness utterance pair.}
    \label{tab:example_testimony}
\end{table*}

We employ two annotators\footnote{All annotators/linguists were compensated in accordance with the institute's norms.\label{lbl:annotate}}, aged 20-30, with advanced language proficiency in English and one senior linguist expert\footref{lbl:annotate}. In the pilot phase, we train our annotators under the supervision of a senior linguist, followed by a consolidation meeting where the disagreements were resolved after deliberation. We repeat the pilot phase for three rounds. Following the pilot phase, our annotators annotate the utterances in isolation and achieve a Cohen's $\kappa = 0.83$ over these labels.   
Finally, our annotators write concise summaries for each attribute conditioned upon the labeled utterances. Each summary is then reviewed and subsequently edited (if necessary) by the senior linguist for relevancy. This two-stage process ensures that the dataset captures rich, contextually relevant interactions and provides human-generated summaries that serve as a gold standard for evaluation. Table \ref{tab:example_testimony} provides an example testimony from MINDSum, illustrating how the utterances of the interrogator and witness are annotated in a testimony. Moreover, it also presents attribute-level summaries for the testimony.

\subsection{Dataset Statistics} Table \ref{tab:counts} presents statistics of the dialogue-act distribution in MINDSum. The dataset comprises 237 testimonies (3,751 utterance pairs between the interrogator and a witness) in the training set, 76 testimonies (1,106 utterance pairs) in the test set, and 63 testimonies (928 utterance pairs) in the validation set, with a 65:20:15 split. This distribution ensures that the model is exposed to a sufficiently large and diverse set of interrogation scenarios during training, while reserving a meaningful portion for robust validation and unbiased testing.
Additionally, label distribution statistics ensure balanced representation across the training, validation, and test splits, thereby enhancing the dataset's reliability for evaluation.
Moreover, we observe that witness utterances are significantly longer than those of interrogators (35.29 vs. 10.99 words in the training set), reflecting the verbosity of testimonies and the need for controlled summarization.

\begin{table}[t]
    \centering
    \renewcommand{\arraystretch}{1.1}
    \resizebox{\textwidth}{!}
    {
    \begin{tabular}{c  c  c  c  c  c  c  c  c  c c c}
        \toprule[1pt]
        & \multicolumn{2}{c}{\bf Avg Uttr. Len} & \multicolumn{3}{c}{\bf Avg Summary Len} & \multicolumn{4}{c}{\bf Utterance Label} & \multirow{2}{*}[1ex]{\bf Num Testimonies} & \multirow{2}{*}[1ex]{\bf Num Utterances} \\ 
        \cmidrule(lr){2-3} \cmidrule(lr){4-6} \cmidrule(lr){7-10} 
        
        \textbf{Split} & \textbf{In} & \textbf{Wt} & \textbf{ED} & \textbf{FD} & \textbf{CD} & \textbf{ED} & \textbf{FD} & \textbf{CD} & \textbf{IF} & & \\
        
        \toprule[1pt]
        
        {\textbf{Train}} & 10.99 & 35.29 & 147.75 & 105.15 & 20.50 & 1860 & 1969 & 117 & 109 & 237 & 3751\\
        
        {\textbf{Test}} & 10.52 & 38.77 & 152.50 & 103.32 & 21.70 & 568 & 564 & 38 & 34 & 76 & 1106\\
        
        {\textbf{Val}} & 10.82 & 38.11 & 143.94 & 110.41 & 20.68 & 458 & 512 & 23 & 31 & 63 & 928\\
        
        \bottomrule[1pt]
    \end{tabular}}
     \caption{Statistics of MINDsum. The train, test, and validation splits are 65:20:15. \textit{Interrogator (In), Witness (Wt), Event Details (ED), Factual Details (FD), Character Description (CD)}, and \textit{Fillers (IF)}.}
    \label{tab:counts}
\end{table}

\section{Method}
Dialogue summarization is a challenging task that requires extracting relevant information from conversational transcripts while preserving context. Conversational data has become increasingly important in NLP, particularly in dialogue systems and human–agent interaction scenarios where contextual understanding is critical \citep{AHMED2024100112}. Moreover, in forensic investigation, information on key aspects of the event, such as \textit{what happened}, \textit{who did it}, \textit{at what time}, etc., is of paramount importance. To ensure all key aspects are adequately captured, we propose CASPER to generate summaries conditioned on specific attributes. 
Formally, we define the task as follows:
Given a dialogue transcript $D = \{u_1, u_2, \dots, u_n\}$ with $n$ pair of utterances, the goal is to generate an attribute-specific summary $S^a$ for a given attribute $a \in \{E, F, C\}$, where $E$, $F$, and $C$ denote \textit{event details, factual details}, and \textit{character descriptions}, respectively.

\begin{figure}[b!]
  \centering
  \includegraphics[width=1.0\textwidth]{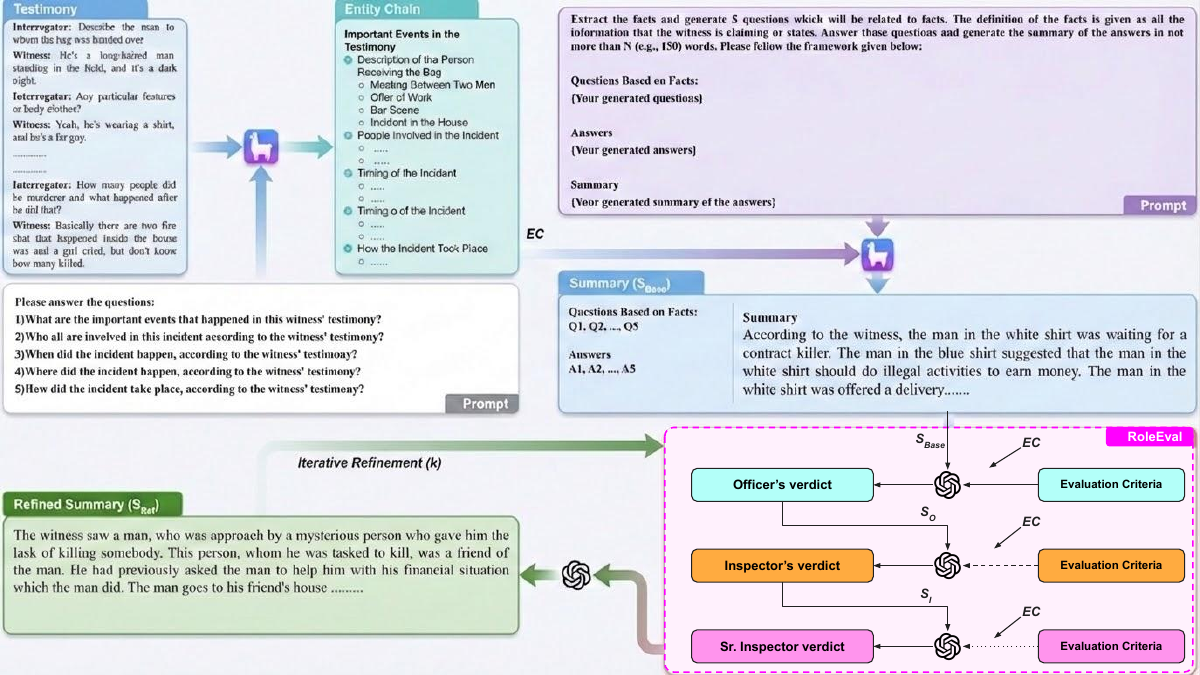}
  \caption{Model architecture of \textbf{CASPER}. The CASPER framework systematically extracts and evaluates attribute-specific information from dialogue transcripts using guiding prompts to generate a summary, and then iteratively refines it to improve the summary. It employs a multiphase assessment mechanism in which language models assume role-specific responsibilities (\textit{officer}, \textit{inspector}, \textit{senior inspector}) to ensure the accuracy, coherence, and completeness of the generated summaries. Through entity extraction, guided questioning, and iterative feedback loops, CASPER produces high-fidelity, attribute-conditioned summaries in a controlled manner.\footref{lbl}}
  \label{fig:model-architecture}
\end{figure}

Our method, shown in Figure~\ref{fig:model-architecture}, integrates large language models (LLMs) with structured guiding prompts, entity extraction, and hierarchical evaluation to achieve high-quality, attribute-focused summaries. To achieve this, CASPER operates in three key stages. 

    \subsection*{\bf Step 1: Dialogue Augmentation with Guiding Prompt.} To direct the language model toward relevant content, we augment the input dialogue $D$ with a structured guiding prompt $P_g$, designed to elicit key aspects of the conversation. The augmented dialogue $D_g$ is then processed by an LLM to extract entities.
    
    \begin{tcolorbox}[float=h, enhanced, breakable, colback=gray!5!white, colframe=black!75!white, 
    title=\textbf{Step 1: Guiding Prompt}, boxrule=0.2pt, arc=4pt, width=\columnwidth, left = 2pt, right = 2pt, top = 2pt, bottom=1pt, boxsep = 2pt, fontupper=\footnotesize]
    \textbf{Please answer the questions:}
    \begin{enumerate}[leftmargin=2em]
        \item \textbf{What are the important events that happened in this witness' testimony?} \\ \textit{{\{\{Your Answer\}\}}}
        
        \item \textbf{Who all are involved in this incident according to the witness' testimony?} \\ \textit{{\{\{Your Answer\}\}}}
        
        \item \textbf{When did the incident happen, according to the witness' testimony?} \\ \textit{{\{\{Your Answer\}\}}}
        
        \item \textbf{Where did the incident happen, according to the witness' testimony?} \\ \textit{{\{\{Your Answer\}\}}}
        
        \item \textbf{How did the incident take place, according to the witness' testimony?} \\ \textit{{\{\{Your Answer\}\}}}
    \end{enumerate}
    \end{tcolorbox}
 
    \subsection*{\bf Step 2: Entity Extraction and Attribute-Specific Summarization.} The LLM (Llama 3.1 - 8B) processes $D_g$, generating a set of entities $E = \{e_1, e_2, \dots, e_k\}$ that serve as the foundation for \textit{attribute-specific prompting}, which generates targeted questions based on the designated attribute $a$. The LLM then responds to these questions, producing attribute-specific answer sets $A^a$, which encapsulate essential details for summarization. The attribute-specific answers $A^a$ are synthesized into a summary $S^a$ of the specific attribute using the Llama 3.1 (8B) model in a zero-shot setting. This summarization step ensures that the generated text aligns with the chosen attribute while maintaining coherence and fluency.

    \begin{tcolorbox}[float=h,enhanced, breakable, colback=gray!5!white, colframe=black!75!white, 
        title=\textbf{Step 2: Attribute-Specific Prompt}, boxrule=0.2pt, arc=4pt, width=\columnwidth, left = 2pt, right = 2pt, top = 2pt, bottom=1pt, boxsep = 2pt, fontupper=\footnotesize]
        \textbf{Extract the \textit{attribute} and generate 5 questions related to \textit{attribute}:} \textit{Definition of the attribute.} \\
        
        \textbf{Answer those questions and generate a summary of the answers in no more than 150 words. Please follow the framework given below:}
        
        \vspace{1ex}
        \textbf{Questions Based on \textit{attribute}:} \textit{Your generated questions} \\
        
        \textbf{Answers:} \textit{Your generated answers} \\
        
        \textbf{Summary:} \textit{Your generated summary of the answers} \\
        \textbf{Note:} Always start the summary with \textit{"According to the witness."}
    \end{tcolorbox}

    \subsection*{\bf Step 3: \textbf{RoleEval} -- Performing Hierarchical Evaluation via Role-Specific Prompts.} \label{Imp_RoleEval} To assess summary quality, we introduce an \textit{hierarchical evaluation framework} using role-specific prompts, \textit{aka.} \textbf{RoleEval}. The role-specific evaluation framework follows a hierarchical structure of \textit{police officer}, \textit{inspector}, and \textit{senior inspector} \citep{agarwal2021exploratory, aristovnik2014performance, brans1985note}\footnote{We selected the roles of Officer, Inspector, and Senior Inspector based on these reference to effectively simulate the real-world responsibilities that come with these positions.} that reflects a structured approach or Promethee method \citep{mareschal2005promethee} to forensic review processes. The \textit{police officer} role primarily focuses on ensuring factual accuracy, whereas the \textit{inspector} takes on a more analytical role by evaluating the investigative relevance of the collected evidence and identifying key elements important for further investigation. Finally, the \textit{senior inspector} addresses the strategic implications of the findings. This role involves synthesizing information gathered by both officers and inspectors, then critically evaluating the summary in terms of its legal and strategic implications. Together, verdicts from these roles create a comprehensive framework that evaluates and helps improve the generated summary. 
        
    \begin{tcolorbox}[float=h,enhanced, breakable, colback=cyan!5!white, boxrule=0.2pt, arc=4pt, left = 10pt, right = 15pt, top = 2pt, bottom=1pt, boxsep = 2pt, fontupper=\footnotesize]
    \begin{itemize}[leftmargin=*]
        \item \textbf{Police Officer:} \textit{Evaluates the summary based on the criteria.}
        \item \textbf{Inspector:} \textit{Reassess the summary generated by police officer and provide a detailed evaluation.}
        \item \textbf{Senior Inspector:} \textit{Final evaluation and verdict based on all evidence.}
    \end{itemize}
    \end{tcolorbox}

    \begin{tcolorbox}[float=b!, enhanced, breakable, colback=gray!5!white, colframe=black!75!white, 
    title=\textbf{Step 3: Role-Specific Verdict Prompt}, boxrule=0.2pt, arc=4pt, width=\columnwidth, left = 2pt, right = 2pt, top = 2pt, bottom=1pt, boxsep = 1pt, fontupper=\footnotesize]

\textbf{Role -- Officer:} \textit{You are a diligent Police Officer responsible for reviewing interrogation summaries. Your role is to ensure that the summary accurately represents the key facts from the original transcript. Focus on whether the summary correctly conveys who was involved, what was said, and the sequence of events. Identify any missing or misleading information and ensure that the evaluation remains clear and easy to understand.} \\

\textbf{Role -- Inspector:} \textit{You are an experienced Inspector overseeing case reviews. Your role is to assess whether the summary not only captures the key facts but also highlights the most relevant information for an ongoing investigation. Evaluate the coherence and logical flow of the summary. Determine if the extracted information is useful for drawing conclusions or further interrogation. Ensure the evaluation is concise but does not omit critical investigative details.} \\

\textbf{Role -- Senior Inspector:} \textit{You are a Senior Inspector with years of experience handling complex cases. Your role is to critically evaluate the summary in terms of its legal and strategic implications. Does the summary provide insights that could guide the next steps in the case? Are there any biases, inconsistencies, or ambiguities in the summary that might impact decision-making? Assess whether the summary maintains objectivity, captures underlying motives or contradictions, and aligns with legal standards for case documentation.}

\tcbline

\textbf{\{Role\}} \\[0.5ex]

Evaluate the summary given below using the evaluation criteria for \textbf{\{Attribute: Factual Details\}} and the context for evaluation. The answer for the questions in the evaluation criteria presented in the evaluation points should only be on a scale of \textbf{1 to 5}, where 1 is the least relevant and 5 is the most pertinent, along with their reasoning. \\[0.3ex]

\vspace{0.5ex}
\textbf{Summary:} \{\textit{Summary}\} \\[0.5ex]

\textbf{Evaluation Criteria:}
\begin{enumerate}[leftmargin=2em, itemsep=0.25ex, topsep=0pt, parsep=0pt]
    \item Coverage – \textit{The extent to which the summary captures all the critical events associated with the attribute from the source dialogue.}
    \item Consistency – \textit{Does the summary maintain logical progression without skipping or misordering key moments? Are any fabricated or incorrect facts included?}
    \item Clarity – \textit{Does the generated summary avoid ambiguity?}
    \item Attribute Relevance – \textit{Are there any unnecessary details present in the generated summary, or do the summaries of one attribute overlap with others?}
    \item Factual Completeness – \textit{Does the generated summary omit any key details / Does the summary capture deceptive elements (such as lies etc.)? Are all critical facts included without omitting any key details?}
\end{enumerate}

\vspace{1em}

\textbf{Context:} \{\textit{Dialogue Data}\} \\[0.4ex]

\textbf{Evaluation Summary:} \{Role's Verdict\}

\end{tcolorbox}
    Each evaluation phase simulates the perspectives of different stakeholders, ensuring a comprehensive assessment. The evaluation criteria, along with the previously generated summary and entity chain, are presented alongside the role-specific prompts to assess the summary and generate verdicts. We use GPT-4o mini for role-specific verdicts. Based on the evaluation verdicts, the summary is subsequently refined. By combining structured prompting, evaluation verdicts, and subsequent refinements, CASPER produces high-quality, attribute-specific summaries that faithfully capture key aspects of the dialogue.

\section{Experiments, Results, and Discussion}
In this section, we present our experimental results and analyses of the MINDSum dataset. 

\subsection{Baselines}
We compare CASPER against multiple PLMs and LLMs on the task of controlled text summarization. 

\begin{itemize}
    \item \textbf{SumCoT} \citep{wang-etal-2023-element} extracts key elements through step-by-step reasoning to enhance factual accuracy. 
    \item \textbf{PromptSum} \citep{ravaut-etal-2023-promptsum} improves controllability via soft and discrete entity prompts to reduce hallucinations. 
    \item \textbf{PLASMA} \citep{naik-etal-2024-perspective} generates perspective-specific summaries using an energy-controlled loss function. 
    \item \textbf{DialoGPT} \citep{zhang-etal-2020-dialogpt} enhances dialogue summarization by extracting keywords, reducing redundancy, and segmenting topics. 
    \item \textbf{Tree of Thoughts (ToT)} \citep{yao2023tree} employs a tree-based search to explore multiple reasoning paths, iteratively pruning less promising branches to optimize decision-making. 
    \item \textbf{Reflexion} \citep{shinn2023reflexion} leverages self-reflection to iteratively refine outputs, using feedback loops to correct errors and improve coherence.
\end{itemize}

Additionally, we include \textbf{GPT-4o-mini} \citep{achiam2023gpt} as an LLM baseline and evaluate fine-tuned \textbf{BART} \citep{lewis-etal-2020-bart}, \textbf{T5} \citep{JMLR:v21:20-074}, and \textbf{PEGASUS} \citep{pmlr-v119-zhang20ae} on MINDsum. PLMs follow Hugging Face’s fine-tuning guidelines, while LLMs are evaluated using CASPER’s attribute-specific zero-shot prompts.

\subsection{Evaluation}
We employ both lexical similarity metrics (ROUGE-1, ROUGE-2, ROUGE-L) \citep{lin-2004-rouge} and semantic similarity metric (BERTScore) \citep{zhang2020bertscoreevaluatingtextgeneration} for the evaluation. Despite their usefulness and popularity for assessing the generative quality, these metrics do not capture the domain-specific nuances important for the evaluation. To support our evaluation, we evaluate our summaries on five dimensions using LLM as a judge -- \textit{coverage}, \textit{consistency}, \textit{clarity}, \textit{attribute relevance}, and \textit{factual completeness}.

\begin{itemize}
    \item \textbf{Coverage:} The extent to which the summary captures all the critical events associated with the attribute from the source dialogue.
    \item \textbf{Consistency:} Does the summary maintain logical progression without skipping or misordering key moments? Are any fabricated or incorrect facts included?
    \item \textbf{Clarity:} Does the generated summary avoid ambiguity?
    \item \textbf{Attribute Relevance:}  Are there any unnecessary details present in the generated summary, or do the summaries of one attribute overlap with others?
    \item \textbf{Factual Completeness:} Does the generated summary omit any key details / Does the summary capture deceptive elements (such as lies etc.)? Are all critical facts included without omitting any key details?
\end{itemize}

We follow and adapt the same hierarchical structure of RoleEval, as outlined in Section \ref{Imp_RoleEval}, to capture role-specific perspectives on these dimensions. In each evaluation phase, we assign a distinct role of \textit{officer}, \textit{inspector}, or \textit{senior inspector} to simulate real-world perspectives and assess the generated summary. RoleEval assigns numerical scores (on a Likert scale of 1 to 5) to summaries based on the predefined evaluation criteria and report the mean of scores from \textit{officer}, \textit{inspector}, and \textit{senior inspector}. These scores provide a structured, interpretable summary of quality, aiding model evaluation and performance benchmarking. 
RoleEval’s multi-tier structure simulates forensic review, assessing information accuracy, investigative relevance, and strategic implications, unlike single-step evaluations focusing solely on lexical (ROUGE) or semantic (BERTScore) overlap.

\subsection{Result Analysis} Table~\ref{tab:performance} presents a comparative analysis of CASPER against baseline models, evaluated using ROUGE (R1, R2, RL), BERTScore (BS), and our RoleEval (RE) framework. CASPER achieves the highest scores in almost all three summarization attributes, significantly outperforming existing techniques and attribute alignment methods. For ED, CASPER achieves a RoleEval score of 86.06, surpassing SumCoT (76.64) and PromptSum (60.04), demonstrating its ability to accurately capture key events. Similarly, CASPER’s RoleEval score of 80.67 for FD outperforms GPT-4o (74.04) and SumCoT (72.18), highlighting its effectiveness in preserving factual information. Furthermore, CASPER attains a RoleEval score of 66.68 in CD, exceeding PromptSum (43.78) and Plasma (28.54) but was outperformed by GPT-4o (69.8) and SumCoT (80.12).

\begin{table*}[t]
\centering
\begin{adjustbox}{width= 1.0\textwidth}
\begin{tabular}{lccccccccccccccccc}
\toprule
& \multicolumn{5}{c}{\bf Event Details} & \multicolumn{5}{c}{\bf Factual Details} & \multicolumn{5}{c}{\bf Character Description} \\
\cmidrule(lr){2-6} \cmidrule(lr){7-11} \cmidrule(lr){12-16}
\multirow{-2}{*}{\bf Model}  & \bf R1 & \bf R2 & \bf RL & \bf BS & \bf RE & \bf R1 & \bf R2 & \bf RL & \bf BS & \bf RE & \bf R1 & \bf R2 & \bf RL & \bf BS & \bf RE\\
\midrule
 PLASMA & 9.83 & 1.87 & 7.81 & 85.71 & 24.91 & 15.27 & 6.48 & 12.99 & 85.71 & 18.39 & 11.18 & 4.34 & 9.95 & 82.22 & 28.54 \\
 T5 & 35.80 & 10.75 & 23.80 & 85.75 & 36.13 & 34.26 & 12.31 & 24.66 & 85.28 & 39.5 & 23.59 & 8.86 & 17.52 & 85.61 & 30.56 \\
 PEGASUS & 38.22 & 12.64 & 25.57 & 86.85 & 52.69 & 31.39 & 12.50 & 21.81 & 86.79 & 67.65 & 38.64 & 17.75 & 29.33 & 88.70 & 45.41 \\ 
 BART & 40.90 & 10.13 & 25.70 & 87.38 & 56.44 & 42.23 & 17.26 & 26.14 & 87.45 & 66.17 & 34.22 & 12.60 & 25.87 & 87.46 & 45.83 \\
  PromptSum & 32.56 & 6.86 & 19.69 & 84.66 & 60.04 & 33.21 & 11.29 & 21.49 & 85.24 & 61.43 & 17.67 & 6.61 & 13.41 & 82.55 &  43.78 \\
  DialoGPT & 32.33 & 9.19 & 22.00 & 87.01 & 63.1 & 33.84 & 9.74 & 22.98 & 87.22 & 70.05 & 32.59 & 9.68 & 22.50 & 87.06 & 46.65 \\
 SumCoT & 40.38 & 12.19 & 26.24 & 88.10 & 76.64 & 41.19 & 15.13 & 20.06 & 88.07 & 72.18 & 30.55 & 9.90 & 23.67 & 86.50 & \textbf{80.12}\textsuperscript{\dag}  \\
 GPT-4o & 42.38 & 12.40 & 25.84 & 88.17 & 80.78 & 41.02 & 13.82 & 26.76 & 88.06 & 74.04 & 24.78 & 11.46 & 19.23 & 87.27 & 69.80 \\
 ToT & 39.15 & \textbf{14.60}\textsuperscript{\dag} & 25.80 & 88.31 & 85.68 & 42.27 & 15.66 & 27.47 & 86.12 & 75.98 & 29.39 & 12.59 & 18.14 & 86.98 & 69.92 \\
 Reflexion & 43.80 & 13.32 & 25.68 & 87.24 & 86.04 & \textbf{44.43}\textsuperscript{\dag} & 14.16 & 26.38 & 86.25 & 78.68 & 27.39 & 11.72 & 23.14 & 87.29 & 69.12 \\
\midrule
    \rowcolor{white}
    \par
 \textbf{CASPER ($k$=1)} & \textbf{44.94}\textsuperscript{\dag} & 14.09 & \textbf{28.54}\textsuperscript{\dag} & \textbf{88.32}\textsuperscript{\dag} & 86.06 & 42.87 & \textbf{16.27}\textsuperscript{\dag} & \textbf{27.59}\textsuperscript{\dag} & \textbf{88.31}\textsuperscript{\dag} & 81.62 & \textbf{40.46}\textsuperscript{\dag} & \textbf{20.44}\textsuperscript{\dag} & \textbf{32.32}\textsuperscript{\dag} & \textbf{89.06}\textsuperscript{\dag} & 66.68 \\
 \midrule
 IR ($k$ = 2) & 43.99 & 13.67 & 27.02 & 88.30 & 84.98 & 40.53 & 13.20 & 24.75 & 87.47 & 80.67 & 27.35 & 7.59 & 18.94 & 86.56 & 66.68 \\
 IR ($k$ = 3) & 42.58 & 13.00 & 26.28 & 88.10 & 84.56 & 38.54 & 11.75 & 23.39 & 87.22 & 79.75 & 25.30 & 6.61 & 17.64 & 86.59 & 65.15 \\
 w/o Labels (\textbf{L}) & 37.80 & 9.37 & 22.68 & 87.04 & \textbf{86.54}\textsuperscript{\dag} & 42.20 & 14.46 & 26.20 & 87.92 & \textbf{82.67}\textsuperscript{\dag} & 27.35 & 7.59 & 18.94 & 86.56 & 67.98 \\
 w/o {IR} & 42.81 & 12.75 & 27.07 & 86.26 & 74.25 & 37.81 & 11.12 & 23.07 & 87.12 & 69.17 & 29.20 & 9.23 & 20.55 & 87.37 & 61.15 \\
 w/o \textbf{L, IR} & 39.56 & 10.47 & 25.15 & 87.43 & 77.2 & 38.62 & 11.40 & 22.79 & 87.07 & 72.48 & 26.51 & 8.33 & 19.19 & 86.05 & 61.2 \\
\bottomrule
\end{tabular}
\end{adjustbox}
\caption{Performance comparison of Baseline Models and CASPER on MINDsum Dataset. \textbf{IR:} Iterative Refinement; \textbf{RE:} RoleEval.}
\label{tab:performance}
\end{table*}

To assess CASPER’s key components, we performed an ablation study. Removing attribute-specific labels and iterative refinement caused a sharp drop of 8.86 (ED), 11.5 (FD), and 5.48 (CD) in RoleEval scores, confirming their importance. Eliminating only iterative refinement further reduced factual consistency, lowering RoleEval scores by 11.81 (ED), 11.5 (FD), and 5.53 (CD), underscoring its role in improving summaries. Interestingly, removing attribute-specific labels alone slightly boosted performance, likely due to iterative refinement's continuous enhancement. To ensure the effectiveness of the entity chains generated from the guiding prompt, we conducted an experiment comparing the entity chains produced by the dialogue to those found in the gold summary. We evaluated the lexical overlap between the two sets using the ROUGE score. This experiment involved 120 samples and resulted in a ROUGE score of 0.8846. This indicates that our extracted entity chains are relevant to attribute-based summarization 88.46\% of the time. 

\paragraph{Statistical Significance:} \label{sec: stats}
A paired two-tailed t-test was conducted on a dataset comprising 120 samples to evaluate the performance improvements of CASPER. The results of the analysis, presented in Table~\ref{tab:significance}, demonstrate that the enhancements observed are statistically significant. This indicates that the likelihood of these improvements occurring by random chance is very low, thereby reinforcing the effectiveness of CASPER in its specific application.

\begin{table}[t]
\centering
{%
\begin{tabular}{lccc}
\toprule
\textbf{Comparison} & \textbf{Mean $\Delta$} & \textbf{$t$} & \textbf{$p$} \\
\midrule
CASPER vs.\ ToT & +6.1 & 3.42 & 0.001 \\
CASPER vs.\ GPT-4o & +5.6 & 2.77 & 0.006 \\
CASPER vs.\ Reflexion & +3.9 & 2.31 & 0.022 \\
\bottomrule
\end{tabular}
}
\caption{Significance of CASPER's improvements.}
\label{tab:significance}
\end{table}

\subsection{Ablation Study} 
To assess the individual contributions of various components in our proposed framework, we perform a detailed ablation study. This analysis enables us to assess the impact of specific design choices on overall performance.

\subsubsection{Hierarchical vs Non-Hierarchical Evaluation}
Table \ref{tab:step_analysis} provides a comprehensive overview of the results from our hierarchical (multi-step) vs non-hierarchical (single-step) evaluation using the RoleEval framework. Our analysis indicates a significant difference in the quality of summaries produced by the different evaluation setups. Specifically, we found that summaries derived from a multi-step evaluation process were more informative and nuanced than those from a single-step approach. This suggests that the multi-step methodology not only captures a broader context but also enriches the information presented in the summaries.

\begin{table}[hb]
\centering
{%
\begin{tabular}{l c c c}
\toprule
\textbf{} & \textbf{ED} & \textbf{FD} & \textbf{CD} \\
\midrule
\textbf{Multi-step (Officer $\rightarrow$ Inspector $\rightarrow$ Senior Inspector)} & \textbf{86.3} & \textbf{82.2} & \textbf{67.0} \\
\midrule
Single-step (Officer) & 77.3 & 74.9 & 61.0 \\
Single-step (Inspector) & 79.2 & 77.7 & 64.4 \\
Single-step (Senior Inspector) & 84.0 & 81.0 & 66.0 \\
\bottomrule
\end{tabular}
}
\caption{Single-step vs multi-step evaluation with RoleEval}
\label{tab:step_analysis}
\end{table}

\subsubsection{Hierarchical Evaluation vs Case Severity}
We examine the effectiveness of our hierarchical evaluation strategy, RoleEval, under varying levels of case severity to understand its robustness across different forensic contexts. We conducted an in-depth analysis by categorizing the dataset into three distinct severity levels: \textbf{\em low}, \textbf{\em mid}, and \textbf{\em high}, based on the criticality of the events described in the interrogations. We list the category-wise cases in Table~\ref{tab:theme_severity}.

\begin{itemize} [noitemsep]
    \item \textbf{Low Severity}: Content with minimal to no depiction of criminal activity, violence, or thematically sensitive material that would be distressing to the average viewer. E.g., \textit{chain snatching}, \textit{petty thefts}, etc.
    \item \textbf{Mid Severity}: Content that depicts clear criminal acts, significant peril, or thematically sensitive situations that could be moderately distressing, but may not feature explicit or graphic life-ending violence. E.g., \textit{self-harm}, \textit{bribery}, etc.
    \item \textbf{High Severity}:  Content featuring explicit, intense, or graphic depictions of life-threatening violence, extreme psychological trauma, or the most serious criminal offenses. E.g., \textit{murder}, \textit{trafficking}, \textit{gangsterism}, etc.
\end{itemize}

\begin{table}[t]
\centering
{%
    \begin{tabular}{l l}
    \toprule
    \textbf{Label} & \textbf{Theme(s)} \\
    \midrule
    \textbf{Low} & Non-Crime, Chain Snatching, Petty Thefts, Phishing \\
    {\textbf{Mid}} & Fraud, Ransom, Robbery, Self-Harm,  Racketeering, Bribery, Fight, Forgery, Bullying \\
   {\textbf{High}} & Murder, Gangsterism, Kidnap, Sexual Assault, Homicide, Drug Trafficking, Accident\\
    \bottomrule
    \end{tabular}%
}
\caption{Event themes grouped by severity label}
\label{tab:theme_severity}
\end{table}

\begin{table}[t]
\centering
{%
\begin{tabular}{l c c c}
\toprule
\textbf{} & \textbf{Low} & \textbf{Mid} & \textbf{High} \\
\midrule
Officer & 84.0 & 80.2 & 76.8 \\
Inspector (Officer $\rightarrow$ Inspector) & 84.9 & 84.2 & 79.2 \\
Senior Inspector (Officer $\rightarrow$ Inspector $\rightarrow$ Senior Inspector) & 88.4 & 85.0 & 80.0 \\
\bottomrule
\end{tabular}
}
\caption{Performance scores across hierarchical roles}
\label{tab:Ablation}
\end{table}

Table \ref{tab:Ablation} shows how the roles in RoleEval perform in such cases. We conduct a study across 120 samples and calculate the average RoleEval score across all three attribute summaries (ED, FD, and CD). We perform experiments in multiple configurations. We observe that, in low-severity cases, all roles performed considerably well, possibly due to the simplicity of the cases, whereas in mild-severity instances, the \textit{police officer} role sometimes missed key details which are eventually captured by \textit{inspector} and \textit{senior inspector}. Similarly, in high-severity cases, the \textit{senior inspector} consistently extracted more critical forensic elements than both \textit{police officer} and \textit{inspector}.

Based on our observation, we argue that in a resource-constrained setting, low-severity events can use the \textit{officer} role to refine the summary, bypassing the experiences of \textit{inspector} and \textit{senior inspector} roles. For mid-level severity crimes, both the \textit{officer} and \textit{inspector} roles can be used for summary refinement, and the \textit{senior inspector} role may be skipped. However, for high-severity cases, it is optimal to use all three roles \textit{officer}, \textit{inspector}, and \textit{senior inspector}—to refine the summary. These findings reinforce the need for hierarchical evaluation mechanisms, such as RoleEval, to ensure accurate forensic summarization across varying case complexities.

\subsubsection{Iterative Refinement} 
We elaborate on the iterative refinement process where the hierarchical evaluation is repeated for $K$ iterations in search for further refinement. 
At each iteration $t$, we integrate feedback to improve $S^a$, ensuring greater fidelity to the dialogue and to attribute-specific constraints. We observe a performance drop with refining the summary for $K >= 2$ (c.f. Table \ref{tab:performance} IR ($K=2$, $K=3$)) due to a decrease in verbosity and including non-essential or irrelevant details. This could be due to the use of comprehensive models like GPT-4o and LLaMA 3.1 in CASPER, which are capable of comprehending essential and fine-grained details in the testimony in a single iteration. 

\begin{table}[!t]
    \centering
    \renewcommand{\arraystretch}{1.5}
    \resizebox{\textwidth}{!}{
    \begin{tabular}{p{5cm} p{5cm} p{5cm}}
    \toprule
         
     \rowcolor{white} \multicolumn{3}{p{16cm}}{\textbf{Reference Summary:} The witness describes a woman named Swati, a receptionist in a hospital, who gets a call from another woman named Rupal, claiming to be the wife of an employee whose name was Vikash. The woman asked for her husband's phone number, but the receptionist refused, citing the hospital's policy against sharing an employee's number. The next day, the woman calls the receptionist again and claims that her husband has been missing for 2 months. The receptionist, feeling sympathetic, agrees to help her. When the receptionist and her senior informed the employee of the incident involving the woman who claimed to be his wife, he denied it, as he was not married, which left them confused. Later, the Witness claims that the woman was not the employee's wife and was making prank calls to extract money from people. \newline \textcolor{green!40!black}{\textbf{Severity:} Low}; \textbf{Attribute:} Event} \\ \hdashline
     \rowcolor{white} \multicolumn{3}{p{16cm}}{\textbf{Initial Summary ($S_{base}$):} According to the witness, Swati received a call from Rupal, who claimed to be Vikas's wife. Swati denied sharing Vikas's contact number due to the hospital's policy. Rupal called again, claiming her husband had been missing for two months. Swati called Vikas, who denied knowing Rupal and stated he was not married. It was later revealed that Rupal's calls were prank calls to extract money. } \\ \hdashline

     \rowcolor{cyan!20!white} \multicolumn{3}{c}{\bf Role's Verdicts} \\ \hdashline
     \rowcolor{white} \textbf{RoleEval(Officer, $S_{base}$) $\quad \Rightarrow \quad V_O$ } & \textbf{RoleEval(Inspector, $S_{O}$) $\quad \Rightarrow \quad V_I$ } & 
     \textbf{RoleEval(Sr. Inspector, $S_{I}$) $\quad \Rightarrow \quad V_S$ } \\ \cmidrule(lr){1-1} \cmidrule(lr){2-2} \cmidrule(lr){3-3}

     \rowcolor{white} \begin{itemize}[leftmargin=*,noitemsep,nolistsep]
        \item[\textcolor{green!40!black}{+}] \textcolor{green!40!black}{Captures key events involving Rupal's actions, Swati's reactions, and the hospital revelations.}
        \item[\textcolor{green!40!black}{+}] \textcolor{green!40!black}{Maintains a clear chronological progression from the initial call to the prank's discovery.}
        \item[\textcolor{green!40!black}{+}] \textcolor{green!40!black}{Preserves essential context and aligns well with the annotated source dialogue.}
        \item[\textcolor{red!80!black}{-}] \textcolor{red!80!black}{Missing details about the identities of Swati, Rupal, and Vikas.}
        \item[\textcolor{red!80!black}{-}] \textcolor{red!80!black}{Minor gaps in contextual grounding of participants.}
    \end{itemize} &
     
    \begin{itemize}[leftmargin=*, noitemsep, nolistsep]
        \item[\textcolor{green!40!black}{+}] \textcolor{green!40!black}{Captures all critical events, including Rupal's actions and Swati's responses.}
        \item[\textcolor{green!40!black}{+}] \textcolor{green!40!black}{Maintains precise chronological order and logical progression.}
        \item[\textcolor{green!40!black}{+}] \textcolor{green!40!black}{Aligns closely with factual content from the source dialogue.}
        \item[\textcolor{red!80!black}{-}] \textcolor{red!80!black}{Contains repetitive phrasing affecting readability.}
        \item[\textcolor{red!80!black}{-}] \textcolor{red!80!black}{Minor stylistic issues in narrative flow.}
    \end{itemize} & 
      
     \begin{itemize}[leftmargin=*, noitemsep, nolistsep]
         \item[\textcolor{green!40!black}{+}] \textcolor{green!40!black}{Captures all key interactions, including Rupal's prank, Swati's responses, and Vikas' involvement.}
        \item[\textcolor{green!40!black}{+}] \textcolor{green!40!black}{Preserves chronological structure and accurately reflects the dialogue timeline.}
        \item[\textcolor{green!40!black}{+}] \textcolor{green!40!black}{Maintains strong alignment with the original dialogues and retains critical context.}
        \item[\textcolor{red!80!black}{-}] \textcolor{red!80!black}{Minor redundancies and repetitive phrasing.}
        \item[\textcolor{red!80!black}{-}] \textcolor{red!80!black}{Scope for improving conciseness and narrative efficiency.}
    \end{itemize} \\ \cmidrule(lr){1-1} \cmidrule(lr){2-2} \cmidrule(lr){3-3}
    
    \rowcolor{white} \multicolumn{1}{l}{\hfill \textbf{CASPER}($S_{base}$, $V_O$) $\quad \Rightarrow \quad S_O$} &
    \multicolumn{1}{l}{\hfill \textbf{CASPER}($S_{base}$, $V_I$) $\quad \Rightarrow \quad S_I$} &
    \multicolumn{1}{l}{\hfill \textbf{CASPER}($S_{base}$, $V_S$) $\quad \Rightarrow \quad S_{Ref}$} \\ \hdashline
     
     \rowcolor{white} \multicolumn{3}{p{16cm}}{\textbf{Refined Summary ($S_{Ref}$):} According to the witness, the \textcolor{blue!80!black}{incident involved a receptionist named Swati who works at a hospital}. She received a \textcolor{blue!80!black}{call from a woman named Rupal, who claimed to be the wife of an employee named Vikash}. The woman requested her husband’s phone number, but the receptionist declined, \textcolor{blue!80!black}{citing the hospital’s policy prohibiting the sharing of employee contact information}. The following day, the same woman called the receptionist again and stated that her husband had been missing for two months. Feeling a sense of compassion, the receptionist decided to assist her. \textcolor{blue!80!black}{After discussing the situation with her supervisor, the employee was informed about the woman who claimed to be his spouse, but he refuted this, saying he was not married, leaving them puzzled}. Eventually, \textcolor{blue!80!black}{the witness claimed that the woman was not the employee's spouse and was making prank calls to extract money from people.}} \\
    
    \bottomrule
         
    \end{tabular}}
    \caption{Assessment of Role's verdict for Low severity type events. \textcolor{blue!80!black}{Blue} text in the refined summary represents information that is refined as per the role's verdicts from the initial summary. \textcolor{green!40!black}{Green} indicates the information from the previous role's verdict that the current role is also addressing in its own verdict. \textcolor{red!80!black}{Red} signifies the additional suggested refinement of the summary by the current role's verdict.}
    \label{tab:role-verdicts-low}
\end{table}

\subsection{Verdict Assessment for Different Roles}
\label{sec:verdict}

The inclusion of role-based verdicts serves as a qualitative evaluation mechanism for assessing the effectiveness of attribute-specific summaries. Rather than functioning as passive observations, these verdicts act as a structured, multi-tiered feedback loop that highlights the gap between raw information extraction and coherent narrative synthesis. By employing personas with varying levels of investigative authority, we simulate a professional review hierarchy in which each role evaluates the summary according to different expectations of informational completeness and interpretive depth. This hierarchical perspective enables the evaluation process to assess whether a generated summary is sufficiently robust for diverse real-world investigative and analytical contexts. We present one instance each for low, mid, and high severity cases in Tables \ref{tab:role-verdicts-low}, \ref{tab:role-verdicts-mild}, and \ref{tab:role-verdicts-high}, repectively.

\paragraph{\bf Low severity case} For instance, Table~\ref{tab:role-verdicts-low} illustrates the verdict progression for a low-severity scenario involving a hospital prank call. In this case, the \textit{officer} successfully captures the primary sequence of events, including the receptionist receiving repeated calls from Rupal and the subsequent realization that the calls were a prank. However, the \textit{officer} also notes that certain contextual details—specifically the identities and roles of Swati, Rupal, and Vikas—are missing from the summary. The \textit{inspector} and \textit{senior inspector} largely agree with the \textit{officer}'s assessment but further emphasize the need for improved readability and clearer narrative structure. This demonstrates how higher-level roles refine the evaluation by focusing on presentation quality and narrative coherence.

\paragraph{\bf Mild severity case} In mild-severity cases, such as the taxi driver's accident presented in Table~\ref{tab:role-verdicts-mild}, the hierarchical refinement becomes more pronounced. Here, the \textit{officer} correctly identifies the key factual events, including the accident and the exchange involving the watch. However, the \textit{officer}'s evaluation overlooks an important contextual detail—the passenger's urgency to attend a friend's wedding. The \textit{inspector} identifies this omission and emphasizes that the summary should explicitly capture this context to preserve the narrative motivation behind the events. Additionally, the \textit{inspector} observes that the \textit{officer}'s verdict references the passenger's kindness, which introduces a subjective interpretation that is not necessary for a strictly event-focused representation. The \textit{senior inspector} further refines this critique by cautioning against drawing assumptions about the passenger's intentions. In this manner, the \textit{officer} provides a basic account of the events, while the higher roles progressively filter out subjective interpretations to maintain forensic neutrality.

\begin{table}[!t]
    \centering
    \renewcommand{\arraystretch}{1.5}
    \resizebox{\textwidth}{!}{
    \begin{tabular}{p{5cm} p{5cm} p{5cm}}
    \toprule

\rowcolor{white} \multicolumn{3}{p{16cm}}{\textbf{Reference Summary:} The witness saw a man who was in a rush to catch his flight to Mumbai and attend his friend's wedding. He booked a cab, and it arrived late and demanded extra money. The man agreed to pay extra and boarded the taxi. The man saw the taxi driver's broken watch and offered his own Samsung watch. After the taxi driver left the man at his destination, the man paid 2000 rupees and insisted on giving the driver his watch, but the driver declined, citing an extra fare. The man then got a message from the taxi's customer care office that the driver had been in a car accident and died, which left him in shock. \newline \textcolor{green!40!black}{\textbf{Severity:} Mild}; \textbf{Attribute:} Event} \\ \hdashline
     \rowcolor{white} \multicolumn{3}{p{16cm}}{\textbf{Initial Summary ($S_{base}$):} According to the witness, the first event was the man taking a taxi ride. During the ride, the taxi driver asked for extra money, and the man gave him his watch in addition to the extra money. The taxi driver's watch was not working, and he had a family with a wife, two children, and a mother. The customer care representative informed the taxi driver that he had been in an accident and had died. The man was going to Mumbai when he took the taxi ride.} \\ \hdashline
     \rowcolor{cyan!20!white} \multicolumn{3}{c}{\bf Role's Verdicts} \\ \hdashline

     \rowcolor{white} \textbf{RoleEval(Officer, $S_{base}$) $\quad \Rightarrow \quad V_O$ } & \textbf{RoleEval(Inspector, $S_{O}$) $\quad \Rightarrow \quad V_I$ } & 
     \textbf{RoleEval(Sr. Inspector, $S_{I}$) $\quad \Rightarrow \quad V_S$ } \\ \cmidrule(lr){1-1} \cmidrule(lr){2-2} \cmidrule(lr){3-3}

    \rowcolor{white}
    \begin{itemize}[leftmargin=*, noitemsep]
    \item[\textcolor{green!40!black}{+}] \textcolor{green!40!black}{Captures key events including the taxi ride, driver's odd behavior, tipping situation, and the accident.}
    \item[\textcolor{green!40!black}{+}] \textcolor{green!40!black}{Maintains logical sequencing of events and highlights the man's urgency due to delay.}
    \item[\textcolor{green!40!black}{+}] \textcolor{green!40!black}{Correctly reflects the driver's refusal of the watch as per the source dialogue.}
    \item[\textcolor{red!80!black}{-}] \textcolor{red!80!black}{Includes references to the man's kindness, which are not directly relevant to event-focused summarization.}
    \item[\textcolor{red!80!black}{-}] \textcolor{red!80!black}{Interpretation of the man's motivations lacks precision.}
    \end{itemize}
    &
    \begin{itemize}[leftmargin=*, noitemsep]
    \item[\textcolor{green!40!black}{+}] \textcolor{green!40!black}{Captures key events including late arrival, driver's unusual behavior, tipping, and the accident.}
    \item[\textcolor{green!40!black}{+}] \textcolor{green!40!black}{Maintains clear chronological order and improves narrative coherence.}
    \item[\textcolor{green!40!black}{+}] \textcolor{green!40!black}{Includes contextual detail of the man's urgency related to a friend's wedding.}
    \item[\textcolor{red!80!black}{-}] \textcolor{red!80!black}{Contains references to the man's kindness that deviate from strict event-focused representation.}
    \item[\textcolor{red!80!black}{-}] \textcolor{red!80!black}{Some interpretations of the man's motivations could be further refined.}
    \end{itemize}

    &
    \begin{itemize}[leftmargin=*, noitemsep]
    \item[\textcolor{green!40!black}{+}] \textcolor{green!40!black}{Captures key events including delayed arrival, driver's behavior, tipping, and the accident.}
    \item[\textcolor{green!40!black}{+}] \textcolor{green!40!black}{Maintains overall logical progression and preserves alignment with the source dialogue.}
    \item[\textcolor{green!40!black}{+}] \textcolor{green!40!black}{Identifies that simplifying certain details can improve clarity.}
    \item[\textcolor{red!80!black}{-}] \textcolor{red!80!black}{Lacks sufficient context regarding the man's urgency.}
    \item[\textcolor{red!80!black}{-}] \textcolor{red!80!black}{Includes references to the man's kindness, detracting from a purely event-based summary.}
    \item[\textcolor{red!80!black}{-}] \textcolor{red!80!black}{Some interpretations of the man's actions may introduce assumptions, affecting accuracy.}
    \end{itemize}\\ \cmidrule(lr){1-1} \cmidrule(lr){2-2} \cmidrule(lr){3-3}

     \rowcolor{white} \multicolumn{1}{l}{\hfill \textbf{CASPER}($S_{base}$, $V_O$) $\quad \Rightarrow \quad S_O$} &
    \multicolumn{1}{l}{\hfill \textbf{CASPER}($S_{base}$, $V_I$) $\quad \Rightarrow \quad S_I$} &
    \multicolumn{1}{l}{\hfill \textbf{CASPER}($S_{base}$, $V_S$) $\quad \Rightarrow \quad S_{Ref}$} \\ \hdashline
    
     \rowcolor{white} \multicolumn{3}{p{16cm}}{\textbf{Refined Summary ($S_{Ref}$):} According to the witness, \textcolor{blue!80!black}{the sequence of events began with the man taking a taxi ride to Mumbai, as he was in a hurry to attend a friend's wedding}. The taxi driver requested extra money for the ride, prompting the man to offer his \textcolor{blue!80!black}{Samsung watch in exchange}. During the ride, they had a discussion about the driver's family, which included \textcolor{blue!80!black}{his wife and two children}. \textcolor{blue!80!black}{Upon arrival at the destination, the man paid 2000 rupees and insisted on giving the driver his watch, but the driver declined, citing an extra fare}. Finally, the man received a call from customer care, informing him that the driver had been in an accident and had died.} \\

    \bottomrule
        
    \end{tabular}}
    \caption{Assessment of Role's verdict for Mild severity type events. \textcolor{blue!80!black}{Blue} text in the refined summary represents information that is refined as per the role's verdicts from the initial summary. \textcolor{green!40!black}{Green} indicates the information from the previous role's verdict that the current role is also addressing in its own verdict. \textcolor{red!80!black}{Red} signifies the additional suggested refinement of the summary by the current role's verdict.}
    \label{tab:role-verdicts-mild}
\end{table}

\begin{table}[!t]
    \centering
    \renewcommand{\arraystretch}{1.5}
    \resizebox{\textwidth}{!}{
    \begin{tabular}{p{5cm} p{5cm} p{5cm}}
    \toprule
      \rowcolor{white} \multicolumn{3}{p{16cm}}{\textbf{Reference Summary:} The witness describes a maid who entered the house of her employers and found out that the master and mistress of the house are both dead. Thinking of it as an opportunity, the maid tried to rob the place. She then discovers that the mistress is still alive and begins to return the stolen goods to their original place. But the mistress, who witnessed all this, called the police and accused the maid of theft and murder of her husband. \newline \textcolor{green!40!black}{\textbf{Severity:} High}; \textbf{Attribute:} Event} \\ \hdashline
     \rowcolor{white} \multicolumn{3}{p{16cm}}{\textbf{Initial Summary ($S_{base}$):} According to the witness, the woman, the mistress of the house, entered the apartment and found the owner and herself unconscious. Lakshmi assumed that both were dead and informed Baban about the incident. Lakshmi planned to rob the house, but the mistress was found alive. The mistress accused Lakshmi of murdering her husband and robbing the house, and Lakshmi was taken into custody by the police. The mistress later confessed to killing her husband and framing Lakshmi for the crime.} \\ \hdashline
     \rowcolor{cyan!20!white} \multicolumn{3}{c}{\bf Role's Verdicts} \\ \hdashline
     
     \rowcolor{white}
      \textbf{RoleEval(Officer, $S_{base}$) $\quad \Rightarrow \quad V_O$ } & \textbf{RoleEval(Inspector, $S_{O}$) $\quad \Rightarrow \quad V_I$ } & 
     \textbf{RoleEval(Sr. Inspector, $S_{I}$) $\quad \Rightarrow \quad V_S$ } \\ \cmidrule(lr){1-1} \cmidrule(lr){2-2} \cmidrule(lr){3-3}
     
    \rowcolor{white}
    
    \begin{itemize}[leftmargin=*, noitemsep]
    \item[\textcolor{green!40!black}{+}] \textcolor{green!40!black}{Captures all relevant factual details about Lakshmi, the victims' condition, and actions taken in the apartment.}
    \item[\textcolor{green!40!black}{+}] \textcolor{green!40!black}{Accurately reflects the source content without introducing incorrect information.}
    \item[\textcolor{green!40!black}{+}] \textcolor{green!40!black}{Maintains a focus on factual assertions while avoiding unnecessary details.}
    \item[\textcolor{red!80!black}{-}] \textcolor{red!80!black}{Some sentence structures reduce clarity and readability.}
    \item[\textcolor{red!80!black}{-}] \textcolor{red!80!black}{Minor omissions in context and motivations limit interpretability.}
    \end{itemize}

    &

    \begin{itemize}[leftmargin=*, noitemsep]
    \item[\textcolor{green!40!black}{+}] \textcolor{green!40!black}{Captures key events including discovery of the bodies, contact with Baban, the robbery plan, and confrontation with the mistress.}
    \item[\textcolor{green!40!black}{+}] \textcolor{green!40!black}{Maintains overall factual alignment with the source dialogue.}
    \item[\textcolor{green!40!black}{+}] \textcolor{green!40!black}{Includes minor character details without detracting from the event-focused narrative.}
    \item[\textcolor{red!80!black}{-}] \textcolor{red!80!black}{Lacks clarity regarding who entered the apartment.}
    \item[\textcolor{red!80!black}{-}] \textcolor{red!80!black}{Could benefit from more concise phrasing to improve readability.}
    \item[\textcolor{red!80!black}{-}] \textcolor{red!80!black}{Some simplifications of the dialogue reduce contextual richness.}
    \end{itemize}

    &
    
    \begin{itemize}[leftmargin=*, noitemsep]
    \item[\textcolor{green!40!black}{+}] \textcolor{green!40!black}{Identifies key elements such as Lakshmi's role and the robbery partnership with Baban.}
    \item[\textcolor{green!40!black}{+}] \textcolor{green!40!black}{Recognizes the need for clearer role attribution and narrative precision.}
    \item[\textcolor{red!80!black}{-}] \textcolor{red!80!black}{Provides insufficient context about Baban and the mistress, limiting narrative completeness.}
    \item[\textcolor{red!80!black}{-}] \textcolor{red!80!black}{Lacks clarity regarding relationships and dynamics between characters.}
    \item[\textcolor{red!80!black}{-}] \textcolor{red!80!black}{Fails to capture Lakshmi's change of heart and subsequent corrective actions.}
    \item[\textcolor{red!80!black}{-}] \textcolor{red!80!black}{Ambiguity remains about who entered the apartment; it should explicitly state that Lakshmi entered.}
    \end{itemize}\\ \cmidrule(lr){1-1} \cmidrule(lr){2-2} \cmidrule(lr){3-3}
    
    \rowcolor{white}
    \multicolumn{1}{l}{\hfill \textbf{CASPER}($S_{base}$, $V_O$) $\quad \Rightarrow \quad S_O$} &
    \multicolumn{1}{l}{\hfill \textbf{CASPER}($S_{base}$, $V_I$) $\quad \Rightarrow \quad S_I$} &
    \multicolumn{1}{l}{\hfill \textbf{CASPER}($S_{base}$, $V_S$) $\quad \Rightarrow \quad S_{Ref}$} \\ \hdashline
    
     \rowcolor{white} \multicolumn{3}{p{16cm}}{\textbf{Refined Summary ($S_{ref}$):} According to the witness, Lakshmi entered the apartment to \textcolor{blue!80!black}{find the owner and the mistress unconscious, initially assuming both were dead}. She called her partner Baban to inform him of the incident, but \textcolor{blue!80!black}{devised a plan to rob the house} instead, thinking it was an opportunity. While ransacking the apartment, \textcolor{blue!80!black}{she discovered the mistress was alive, but just as she attempted to rectify her actions, the mistress confronted her, accusing Lakshmi of murder and robbery.} The police arrived shortly after, summoned by the mistress, who had been feigning death, \textcolor{blue!80!black}{ultimately framing Lakshmi for the crime, leading to Lakshmi's arrest}.} \\

    \bottomrule
         
    \end{tabular}}
    \caption{Assessment of Role's verdict for High severity type events. \textcolor{blue!80!black}{Blue} text in the refined summary represents information that is refined as per the role's verdicts from the initial summary. \textcolor{green!40!black}{Green} indicates the information from the previous role's verdict that the current role is also addressing in its own verdict. \textcolor{red!80!black}{Red} signifies the additional suggested refinement of the summary by the current role's verdict.}
    \label{tab:role-verdicts-high}
\end{table}

\paragraph{\bf High severity case} The hierarchical evaluation becomes even more critical in high-severity cases, such as the murder and robbery incident discussed in Table~\ref{tab:role-verdicts-high}. In this scenario, the \textit{officer} captures the discovery of the bodies and the robbery attempt but fails to identify the specific individual who entered the house and the mistress's attempt to feign death. The \textit{inspector} addresses these omissions by identifying the partnership between Lakshmi and Baban and clarifying the robbery scheme, thereby adding the necessary causal structure to the \textit{officer}'s event-level description. Despite this refinement, the \textit{senior inspector} notes that the summary still lacks clarity regarding Baban's identity and role in the narrative. The \textit{senior inspector}, therefore, emphasizes the importance of explicitly stating that Lakshmi entered the house and highlighting her subsequent change of heart. These details were absent in the original summary and were not fully captured by the earlier role assessments, illustrating the deeper analytical scrutiny applied at higher levels of the evaluation hierarchy.

The significance of this role-based evaluation framework lies in its ability to stress-test generated summaries against varying professional standards of scrutiny. By progressively refining summaries through hierarchical verdicts, the framework transforms summarization from a simple text-compression task into a structured decision-support process. In investigative contexts, seemingly minor narrative elements can significantly influence interpretation. For instance, emphasizing a suspect's ``kindness'' may introduce irrelevant bias, whereas omitting critical interactions—such as the involvement of Baban in a robbery scheme—can result in substantial evidentiary gaps. Through systematic analysis of these role-based verdicts, we demonstrate that the proposed framework evaluates summaries not only for informational completeness but also for their practical utility in relation to the severity and investigative significance of the underlying events.

\subsection{Error Analysis}

We also perform error analysis of the obtained results. Table \ref{tab:model-summaries-remarks} presents a comparative analysis of summaries for the {\textit{event details}} generated by CASPER and other baseline models. Most of the baselines either omit critical information (e.g., PromptSum, SumCoT, GPT-4o) or introduce factual inaccuracies (e.g., Reflexion). Additionally, ToT fails to clarify the origins of events. In contrast, CASPER captures all key events convincingly, but commit a minor mistake of misrepresenting the "feedback" as  "text". These findings highlight CASPER’s relative strength in content coverage despite minor lexical inaccuracies. 

\begin{table*}[t!]
\centering
\renewcommand{\arraystretch}{1.2}
\resizebox{\textwidth}{!}{%
\begin{tabular}{l p{12cm} p{3.2cm}}
\toprule
\rowcolor{white} \textbf{Model} & \textbf{Summary} & \textbf{Remarks} \\
\midrule
\rowcolor{red!20}
\textbf{Ref Summary} & 
The witness saw a delivery boy delivering a package to a woman at her house. After he delivered the package, he received a feedback from the which said, "Please save me. I'm in danger." After some consideration, the delivery boy went back to the house and went inside it. The boy heard a noise coming from the inner room, where he found a man lying in a tub. Then the lady suddenly arrived and said he should not be there and that she was going to kill him. The witness says that she takes a gun but slips and shoots herself. The boy, shocked, leaves the place.  & \\
\midrule

\rowcolor{cyan!20}
\textbf{PromptSum} & 
\textbf{I'm in danger." So after some time, after thinking for some time, he went inside the room} and saw that things were not in their right place. Interrogee: When he entered the house, he saw that things were not in the right place, and after some time, a noise was coming from inside the room. Interrogee: When he entered the room, he saw a man lying in a tub, and he was dead.  &
The summary does not cover all the aspects of the event details and also lacks fluency.\\

\rowcolor{white} \textbf{SumCoT} & 
According to the witness, the delivery boy encountered a series of events at the house, including \textbf{a confrontation with the chef, finding a dead man in the tub, receiving a mysterious message, and witnessing the woman accidentally shoot herself}. Despite the unsettling events, the delivery boy ultimately left the house after the incident. &
The summary is lacking specific details about what actually took place during the event. \\

\rowcolor{cyan!20} \textbf{GPT-4o} & 
According to the witness, the delivery boy had a tense exchange with the chef regarding his lateness and poor ratings. \textbf{After delivering the order}, he received a distress message indicating danger. Upon entering the house, he noticed disarray, which raised his suspicions. In the inner room, he discovered a dead man in a tub. When \textbf{confronted by a woman with a gun}, she accidentally shot herself due to a slip. The sequence of events highlights a dramatic turn from a simple delivery to a life-threatening situation. &
The summary is missing key details, such as the order being placed by a woman and that the same woman confronted him at the end. \\

\rowcolor{white} \textbf{ToT} & 
According to the witness, the delivery boy entered the woman's house after \textbf{ receiving a distress message}. He discovered the dead man in the tub, leading to a \textbf{confrontation with the woman who threatened him}. The situation escalated when she accidentally shot herself while aiming a gun, prompting the delivery boy to leave in shock. &
The summary is unclear about the origin of the distress message and the identity of the woman who threatened him. \\

\rowcolor{cyan!20} \textbf{Reflexion} & 
According to the witness, the delivery boy was delivering a package to a woman, \textbf{who was wearing a red blouse}, before he received the message saying "Please save me. I'm in danger". The woman gave him a tip of 400 dollars and offered him a drink. When the delivery boy entered the house, he found a dead body in the tub. The woman said, "You must not be here now. I will kill you," to the delivery boy, and she accidentally shot herself. The delivery boy left the place after the incident. &
The summary describes the woman, which should be included as CD. It incorrectly interprets the "feedback" as a "message" received by the delivery boy.
\\
\midrule

\rowcolor{green!20}\textbf{CASPER} & 
According to the witness, the delivery boy delivered a package to a woman. After receiving a generous tip of \$400, he was \textbf{startled by a text stating}, "Please save me. I'm in danger." Upon entering the house, he discovered chaos and heard suspicious noises from an inner room. Investigating further, he found a man submerged in the tub, deceased. The woman appeared and warned him, "You must not be here now. I will kill you." During the ensuing chaos, she accidentally shot herself while handling a gun. Distraught and shocked, the delivery boy quickly fled the scene. &
The summary included all the necessary details but with a minor mistake of mentioning "text" instead of "feedback". \\
\bottomrule

\end{tabular}
}
\caption{Comparison of summaries of CASPER with other Baselines}
\label{tab:model-summaries-remarks}
\end{table*}

To quantify these observations, we perform a fine-grained analysis of 120 samples across the \textit{event details}, \textit{factual details}, and \textit{character description} attributes across five different failure types. This approach allows us to distinguish between the breadth of failure (how many summaries are affected) and the depth of specific linguistic or factual inaccuracies. We define each failure type as follows:

\begin{itemize}
    \item Attribute Overlap: Leakage of content from one attribute into another (e.g., CD elements appearing in ED summaries)
    \item Omission of Nuance: Key contextual or causal details missed despite overall coherence.
    \item Hallucinated Causality: Incorrect inference of causal or temporal links not grounded in testimony.
    \item Entity Drift: Inconsistent reference or role confusion across utterances.
    \item Lexical Ambiguity: Vague or repetitive phrasing leading to forensic uncertainty.
    
\end{itemize}

\noindent We then define two fundamental ratios to evaluate model performance:

\begin{enumerate}
    \item \textbf{Summary Error Density ($P_e$):} This represents the proportion of generated summaries that contain at least one instance of an error.
    \begin{equation}
        P_e = \frac{N_{es}}{N_{total}}
    \end{equation}
    where $N_{es}$ is the number of summaries containing at least one error, and $N_{total}$ is the total number of summaries generated in the test set.

    \item \textbf{Error Prevalence ($\omega_i$):} This ratio identifies the prevalence of a specific error type $i$  relative to the total pool of errors identified.
    \begin{equation}
        \omega_i = \frac{E_i}{\sum_{j=1}^{n} E_j}
    \end{equation}
    where $E_i$ is the count of errors of type $i$, and $n$ represents the total number of distinct error categories.
\end{enumerate}

\begin{table}[t]
\centering
{
\begin{tabular}{llcccccc}
\toprule
\textbf{Metric} & \textbf{Error Category} (\textcolor{red}{$\downarrow$}) & \textbf{CASPER} & \textbf{GPT-4o} & \textbf{Reflexion} & \textbf{ToT} & \textbf{SumCoT} & \textbf{PromptSum} \\ \midrule
$P_e$ (in \%) & \textit{Overall Density} & \textbf{0.18} & 0.34 & 0.42 & 0.48 & 0.55 & 0.62 \\ \midrule

& Attribute Overlap & \textbf{0.03} & 0.08 & 0.10 & 0.14 & 0.18 & 0.22 \\
& Omission of Nuance & \textbf{0.05} & 0.09 & 0.12 & 0.11 & 0.15 & 0.19 \\
& Hallucinated Causality & \textbf{0.02} & 0.07 & 0.14 & 0.12 & 0.14 & 0.16 \\
& Entity Drift & \textbf{0.04} & 0.08 & 0.09 & 0.15 & 0.12 & 0.16 \\
\multirow{-5}{*}{$\omega_i$ (in \%)}  & Lexical Ambiguity & \textbf{0.04} & 0.09 & 0.13 & 0.12 & 0.13 & 0.15 \\ \bottomrule
\end{tabular}}
\caption{Granular Error Analysis: Summary Error Density ($P_e$) and Error-Type Prevalence ($\omega_i$) for CASPER vs. Baselines.}
\label{tab:error_metrics}
\end{table}

The results in Table \ref{tab:error_metrics} provide a stark contrast between CASPER and the baselines. Our model achieves the lowest summary error density ($P_e = 0.18$), indicating that more than 80\% of its summaries are entirely free of the five defined failure types. In comparison, PromptSum and SumCoT exhibit $P_e$ values exceeding 0.50, suggesting that more than half of their generated outputs contain at least one forensic flaw. The multi-row analysis in Table \ref{tab:error_metrics} illustrates the specific failure modes of each model. While baseline models exhibit high prevalence across all categories, CASPER maintains consistently low values for each individual $\omega_i$. Specifically, CASPER demonstrates a marked resilience to \textit{hallucinated causality} ($\omega_3 = 0.02$) and \textit{attribute overlap} ($\omega_1 = 0.03$). This suggests that the model effectively decouples different narrative attributes, preventing the leakage of character descriptions into event summaries—a common pitfall for general-purpose LLMs like GPT-4o and PromptSum.

The low prevalence of \textit{entity drift} and \textit{lexical ambiguity} in CASPER signifies its ability to maintain stable references throughout the interrogation timeline. In contrast, models like ToT and Reflexion achieve higher scores on \textit{hallucinated causality}, because their iterative reasoning steps introduce imagined connections not present in the witness testimony. We argue that CASPER’s superiority is rooted in three primary factors:

\begin{enumerate}
    \item \textbf{Resilience to Hallucinated Causality:} While baselines like Reflexion often fill in the gaps of an interrogation to create a smoother narrative, GPT-4o confuses witness inference with established facts. Whereas CASPER’s attribute-specific summarization ensures it strictly adheres to the provided testimony, which results in a significantly lower $\omega_i$.
    
    \item \textbf{Structural Attribute Integrity:} The most common failure in PromptSum and ToT is \textit{attribute overlap}, where character descriptions overlap with event details. CASPER’s architecture enforces a clear boundary between attributes, ensuring that $P_e$ remains low even when the source dialogue is convoluted or lacks clear transitions.

    \item \textbf{Mitigation of Entity Drift:} As seen in the comparison with ToT, CASPER maintains a consistent reference to personas (e.g., "the man in the white shirt") throughout the summary. The lower cumulative $\omega_i$ for CASPER suggests that it handles the "who-did-what" complexity of interrogations with higher clinical precision than general-purpose LLMs.
\end{enumerate}

Ultimately, CASPER achieves a significantly lower $P_e$ and $\omega_i$ by enforcing strict attribute-specific boundaries and minimizing hallucinated causality, thereby ensuring that the majority of generated summaries are free of forensic inaccuracies. This reduction in error prevalence signifies a robust alignment with the high-fidelity demands of investigative roles, proving that CASPER is uniquely capable of maintaining factual integrity where general-purpose baselines succumb to narrative drift.

\subsection{Human Evaluation Protocol}

We conduct human evaluation with thirty graduate-level volunteers (CEFR C1+ proficiency)\footnote{https://en.wikipedia.org/wiki/Common\_European\_Framework\_of\_Reference\_for\_Languages}, drawn from computer science and computational linguistics programs. We employ a strict double-blind setup: evaluators are not informed of model identities, and the coordinator managing assignments likewise has no access to the model conditions. Each evaluator assesses 45 summaries along five dimensions —\textit{coverage}, \textit{consistency}, \textit{clarity}, \textit{attribute relevance}, and \textit{factual completeness}— using a 1–5 Likert scale. To ensure reliability, evaluators must provide brief justifications for their scores, enabling consistent interpretation of the rubric guidelines. We observe high inter-rater reliability with Cronbach’s $\alpha = 0.91$, indicating strong internal consistency across evaluators and attributes. This protocol ensures that the human evaluation aligns with the forensic and attribute-specific needs of MINDSum while providing a robust basis for comparing CASPER against multiple baselines. Table~\ref{tab:human-eval} reports mean Likert scores across models.

\begin{table}[h]
\centering
\setlength{\tabcolsep}{8pt}
{
\begin{tabular}{lccccc}
\toprule
\textbf{Model} & Coverage & Consistency & Clarity & Attribute Relevance & Factual
Completeness \\
\midrule
GPT-4o & 4.2 & 4.1 & 4.8 & 3.7 & 4.1 \\
ToT & 4.5 & 4.3 & 4.6 & 4.2 & 4.3 \\
Reflexion & 4.4 & 4.2 & 4.7 & 4.1 & 4.4 \\
CASPER & \textbf{4.8} & \textbf{4.4} & \textbf{4.8} & \textbf{4.9} & \textbf{4.7} \\
\bottomrule
\end{tabular}}
\caption{Human evaluation results (1–5 scale).}
\label{tab:human-eval}
\end{table}

\subsection{Mitigating bias b/w RoleEval as a validator and RoleEval as a metric.} \label{sec: IR-bias_1}
To mitigate bias introduced by using similar frameworks of \textit{RoleEval} as a validator during the refinement process and later as an evaluation metric, we employ two distinct large language models. Utilizing two LLMs (GPT-4o-mini and Phi 4), we experiment with all four combinations. We observe that RoleEval tends to assign higher scores to the summaries, when the same LLM is employed for refinement. On the other hand, we observe a decrease in the scores when different LLMs are used for refinement and evaluation. This phenomenon supplements the understanding of  LLMs favoring their own responses, as highlighted in earlier research \citep{panickssery2024llm}. Table \ref{tab:model_bias} illustrates this trend, showing that both Phi-4 and GPT-4o-mini give preferential scores to the refined summaries generated by the same model. 

\begin{table}[t!]
\centering
\setlength{\tabcolsep}{8pt}
{%
    \begin{tabular}{llcccc}
    \toprule
    \multirow{2}{*}[0.8ex]{\textbf{Base Summarizer}} & \multirow{2}{*}[0.8ex]{\textbf{RoleEval}} & \multirow{2}{*}[0.8ex]{\textbf{Attributes}} & \multicolumn{3}{c}{\textbf{RoleEval (Metric)}} \\ \cmidrule{4-6}
    &  \bf (Validator) & & \bf Phi 4 & \bf GPT-4o mini & \bf Human Eval \\ \midrule
     &  & ED & 89.63 & 86.80 & 86.64 \\
     & & FD & 85.4 & 83.10 & 82.67 \\
     & \multirow{-3}{*}{Phi 4} & CD & 59.52 & 41.22 & 42.09  \\ \cmidrule{2-6}
    &  & ED & 88.32 & 88.24   & 87.56 \\
     & & FD & 81.62 & 83.11 & 81.02 \\
    \multirow{-6}{*}{\rotatebox{0}{Llama 3.1 (8B)}} & \multirow{-3}{*}{GPT-4o mini} & CD & 66.68 & 72.22 & 65.02 \\ 
     \bottomrule
    \end{tabular}%
}
\caption{RoleEval with different LLMs and Human evaluation.}
\label{tab:model_bias}
\end{table}
Further, to determine which approach is more effective, we conduct a human evaluation of the generated summaries. We randomly select 15 testimonies from our dataset and assess the refined summaries produced by both Phi-4 and GPT-4o mini. The human evaluation scores, presented in Table \ref{tab:model_bias}, are relatively closer when using different LLMs for refinement and evaluation.

\subsection{Reliability of RoleEval as an evaluation metric} \label{sec: IR-bias_2}
Since RoleEval utilizes an LLM as a judge, we assess its reliability as a metric. We compute correlation between RoleEval and the human judgment. Moreover, we also assess the correlation between other automatic metrics against the human judgment. Table~\ref{tab:metric-corr} shows that RoleEval exhibits far stronger correlation with human judgments than ROUGE or BERTScore.

\begin{table}[h]
\centering
{%
\begin{tabular}{llccc}
\toprule
\multicolumn{2}{l}{\textbf{Metric Pair}} & \textbf{Pearson $r$} & \textbf{Spearman $\rho$} & \textbf{$p$-value} \\
\midrule
ROUGE-L & $\leftrightarrow$ \quad Human Mean & 0.42 & 0.39 & 0.018 \\
BERTScore & $\leftrightarrow$ \quad Human Mean & 0.58 & 0.55 & 0.006 \\
RoleEval & $\leftrightarrow$ \quad Human Mean & 0.87 & 0.83 & $<$0.001 \\
\bottomrule
\end{tabular}
}
\caption{Correlation of automatic metrics with human judgment.}
\label{tab:metric-corr}
\end{table}

\subsection{Computation Cost and Inference Latency}
Profiling on an NVIDIA A100 GPU shows a 10–15-second inference time for a 50-utterance dialogue (500–700 words), with 60\% of the time attributed to the iterative refinement stage (three LLM passes for the officer, inspector, and senior inspector roles). The lightweight entity extraction and questioning stages minimize overhead. The main bottleneck is the LLM's token generation during summarization and refinement, which depends on dialogue length and summary complexity. For MINDSum's average dialogue (about 35.29 words for witnesses and 10.99 for interrogators), around 1,500-2,000 tokens are processed per dialogue, including prompts and outputs.

\section{Conclusion}
We present CASPER, a framework for controlled dialogue summarization that leverages \textit{}{chain-of-thought attribute-specific prompting} and iterative evaluation to generate high-quality, interpretable summaries. By conditioning on three key attributes (event details, factual details, and character description) and using {RoleEval} for structured assessment, CASPER improves factual completeness, coherence, and relevance. We also introduce {MINDSum}, a curated dataset for forensic summarization, providing a valuable benchmark for evaluating attribute-aligned summaries. Experimental results demonstrate CASPER’s effectiveness, outperforming baselines in automatic and human evaluations. While CASPER demonstrates notable performance, its functionality is currently limited to a monolingual English context. To enhance its applicability and broaden its impact, future research should prioritize expanding CASPER to support multilingual settings.

\section{Ethical Considerations}

The development of the \textbf{MINDSum} dataset for dialogue summarization in forensic settings requires careful ethical considerations. Our research emphasizes data privacy, consent, fairness, and responsible AI use to manage risks associated with sensitive data. We utilized multiple annotators with high inter-annotator agreement (Cohen's $\kappa = 0.83$) to minimize biases, ensuring diverse interrogation scenarios without reinforcing harmful stereotypes. The dataset avoids content that disproportionately impacts specific demographics, focusing on neutrality and factual accuracy. We discourage the misuse of MINDSum to prevent biased decision-making in criminal investigations. This dataset supports the development of transparent and fair summarization techniques. Researchers are urged to adhere to ethical AI principles, aiming for transparency and accountability while protecting participant rights and promoting socially beneficial applications.


\defbibnote{preamble}{}

\printbibliography[prenote={preamble}]

\end{document}